\RequirePackage{fix-cm}
\documentclass[twocolumn]{svjour3}

\usepackage[utf8]{inputenc}
\usepackage{longtable,multirow,supertabular,afterpage}
\usepackage{amssymb,amsbsy,textcomp,marvosym,caption,threeparttable}
\usepackage{mathptmx, amsmath, amsfonts}
\usepackage{bbm} 

\usepackage{nicefrac,xfrac}
\usepackage{eurosym,fancyhdr,CJK,multicol,graphics,indentfirst,color,bm,upgreek,booktabs,graphicx,subfigure}
\usepackage[mathcal]{euscript}
\usepackage{setspace}
\usepackage{float}
\usepackage{hyperref}

\usepackage{array}
\usepackage{enumitem}
\usepackage[backend=bibtex,style = nature,sorting=none]{biblatex}
\usepackage{makecell}

\newcolumntype{P}[1]{>{\centering\arraybackslash}p{#1}} 

\begin{document}\sloppy
	\title{\centering A horizon line annotation tool for streamlining autonomous sea navigation experiments}
	\author{\centering
		Yassir ZARDOUA$^1$ \and Abdelhamid EL WAHABI \and Mohammed BOULAALA$^1$\\\and Abdelali ASTITO$^1$ \and Mhamed EL MRABET}
	
	\institute{Yassir Zardoua \textbf{(Corresponding author)}\at
		\email{yassirzardoua@gmail.com}           
		\and
		Boulaala Mohammed \at
		\email{m.boulaala@gmail.com}
		\and
		Abdelhamid EL WAHABI \at
		\email{elwahabi.abdelhamid@gmail.com}\\
		Astito Abdelali \at
		\email{abdelali\_astito@yahoo.com}\\
		\\
		$^1$Laboratory of Informatics, Systems \& Telecommunications, FSTT, Abdelmalek-Essaadi University, Tetouan, Morocco
	}

	\maketitle

\begin{abstract}
Horizon line (or sea line) detection (HLD) is a critical component in multiple marine autonomous navigation tasks, such as identifying the navigation area (i.e., the sea), obstacle detection and geo-localization, and digital video stabilization. A recent survey highlighted several weaknesses of such detectors, particularly on sea conditions lacking from the most extensive dataset currently used by HLD researchers. Experimental validation of more robust HLDs involves collecting an extensive set of these lacking sea conditions and annotating each collected image with the correct position and orientation of the horizon line. The annotation task is daunting without a proper tool. Therefore, we present the first public annotation software with tailored features to make the sea line annotation process fast and easy. The software is available at: \url{https://drive.google.com/drive/folders/1c0ZmvYDckuQCPIWfh_70P7E1A_DWlIvF?usp=sharing}.


\end{abstract}


%

\section{Introduction}
\subsection{Definition of the horizon line}
\label{intro}
\begin{figure*}[h!]
	\centering
	\subfigure[]{
		\includegraphics[width = 0.47\linewidth, keepaspectratio]{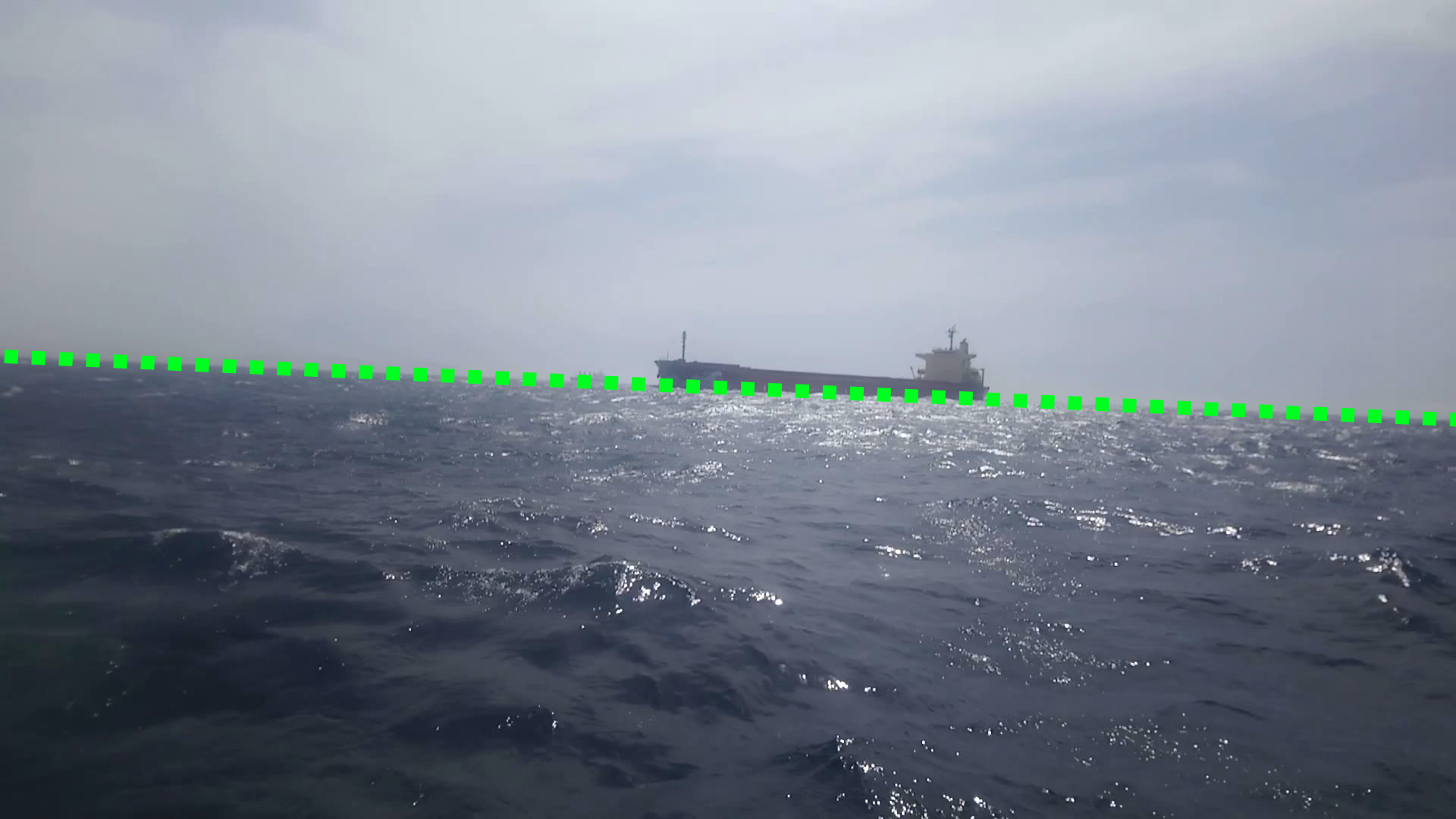}
		\label{fig_sea_horizon}}
	\subfigure[]{
		\includegraphics[width = 0.47\linewidth, keepaspectratio]{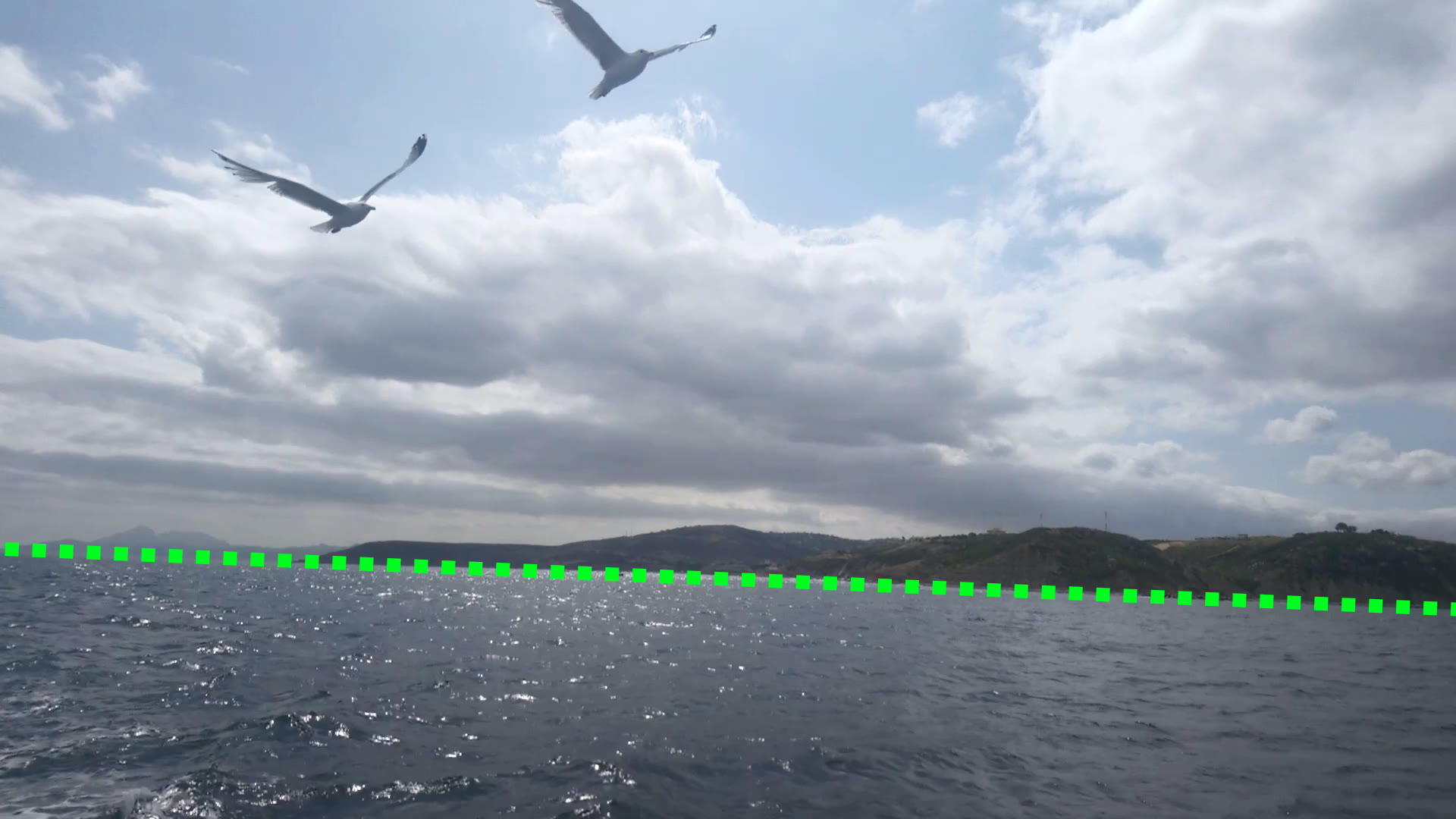}
		\label{fig_coastal_horizon}}
	
	\subfigure[]{
		\includegraphics[width = 0.5\linewidth, keepaspectratio]{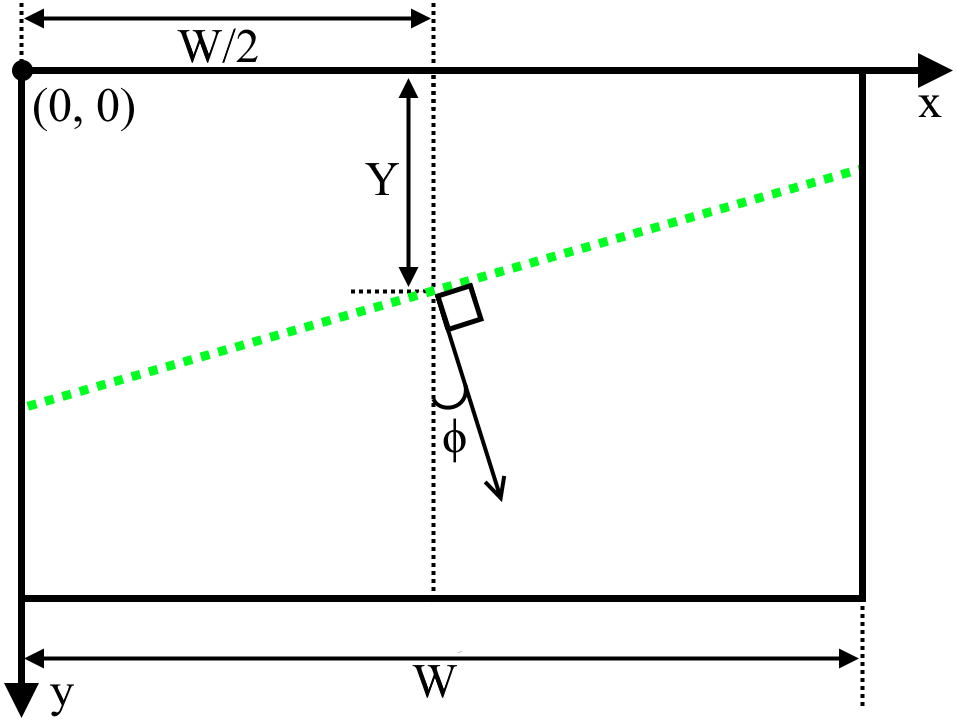}
		\label{fig_horizon_rep}}
	\caption{Definition and representation of the horizon: the horizon line separates the sea from the sky (a) and the sea from the coast (b); (c) $\{Y, \phi\}$ representation on Cartesian coordinates $xy$ of images}
	\label{fig_horizon_examples_and_representation}
\end{figure*}

In maritime images captured from terrestrial platforms, such as buoys, ships, and USVs, the horizon feature (sometimes called the sea-sky line) is defined as the line separating the sea region and the region right above it~\cite{zardoua2021survey} (see Figure~\ref{fig_sea_horizon} and~\ref{fig_coastal_horizon}). The literature includes several ways to represent the location of the horizon line. The position and tilt representation, which we denote by the pair of parameters $\{Y, \phi\}$ (see Figure~\ref{fig_horizon_rep}), is by far the most common and useful representation~\cite{zardoua2021survey, EOsurvey2017} because it provide a direct and precise measure of the horizon's location. Additionally, measuring the detection error based on this representation provides a clear and intuitive idea of how well the horizon detection algorithm is performing.

\subsection{Applications of the horizon line}
The published literature shows that the sea horizon line exhibits a wide range of applications, especially in terms of autonomous navigation. In the case of autonomous sea navigation, the horizon feature is involved on almost every stage of the processing pipeline. We'll cite four major application categories. First, numerous papers and patents used the horizon line to auto-calibrate the camera~\cite{rs13142795,Shin2018, assadzadeh2021automatic}, which include non-maritime scenes as well~\cite{calibration, calibration2}. Second, the position and tilt of the horizon allows the computation of rotational and translational matrices that would digitally stabilize captured video frames~\cite{overview_stabilization, cai, fgsl, 9728058, smtj2015, liu2014, yao1995electronic, duric1996image, Reichert2021}. A stabilized video is not only  easier to watch but facilitates subsequent analyses as well~\cite{Reichert2021, EOsurvey2017, fgsl}. Third, many works used the horizon to segment the image (into sea and non-sea region). In the same context, the horizon feature provides computational benefits to the object detection task by reducing the search region down to that around the horizon or by eliminating the sky pixels that will obviously not depict any obstacle~\cite{asv, Qiao2022, KONG2016185, bai2021ship, LeeVR, s19184004, 7237862, Kocak2012, ZHANG201753,jmse9121408, jmse8100799, 7984596, 9309222, ying2008algorithm, 8908899, ZHANG20154475, 8740763, Shin2018, liu2021real}. Fourth, even after the detection of maritime objects on the image coordinates, some researchers used the horizon line to geo-localize surrounding objects and obstacles~\cite{6070512, Brejcha2017, land_detection, distanceestimation2}, i.e., find the distance in meters from the camera to the obstacle.

\subsection{The need of custom horizon annotation tool}
Our decision to develop and publicly share our annotation tool was driven by the need for a fast and reliable sea line annotation tool, which we clarify in the following points:\\
\textbullet~~the most extensive dataset, the SMD, lacks various sea conditions that must be collected and properly annotated\\
\textbullet~~the sea line annotation of the SMD include significant mistakes and must be corrected before the experiment\\
\textbullet~~many papers such as~\cite{liang, li2021sea, fgsl, smtj2015} include their own private image dataset but do neither mention how did they annotate their image nor share their annotation tool.\\
\textbullet~~currently available tools, such as \textit{LabelMe} and \textit{SuperAnnotate}, lack public free access and customization. The proposed software freely provide unique and tailored features and can be directly used without further modifications.

Section~\ref{sec_app_description} outlines the primary components and features of the proposed software, which are summarized below:
\noindent
\textbullet~~An intuitive graphical user interface (GUI) and easy browsing through video frames\\
\textbullet~~Visualization tools, such as the ability to adjust the thickness of the annotated sea line\\
\textbullet~~The option to duplicate current annotations onto previous non-annotated frames, which is particularly useful for on-shore fixed cameras\\
\textbullet~~Shortcut keys and the ability to annotate all video frames with just the mouse\\
\textbullet~~Misuse warnings, such as alerting the users if they forget to annotate all the frames\\
\textbullet~~The ability to save incomplete annotation files and return to them later for completion, which is especially helpful for video files with a large number of frames.\\

Overall, these features enhance the functionality and user-friendliness of the software and make it a valuable tool for video annotation.

\section{Description of the application: interface blocks and features}
\label{sec_app_description}
\begin{figure*}[h!]
	\includegraphics[width = 1\linewidth, keepaspectratio]{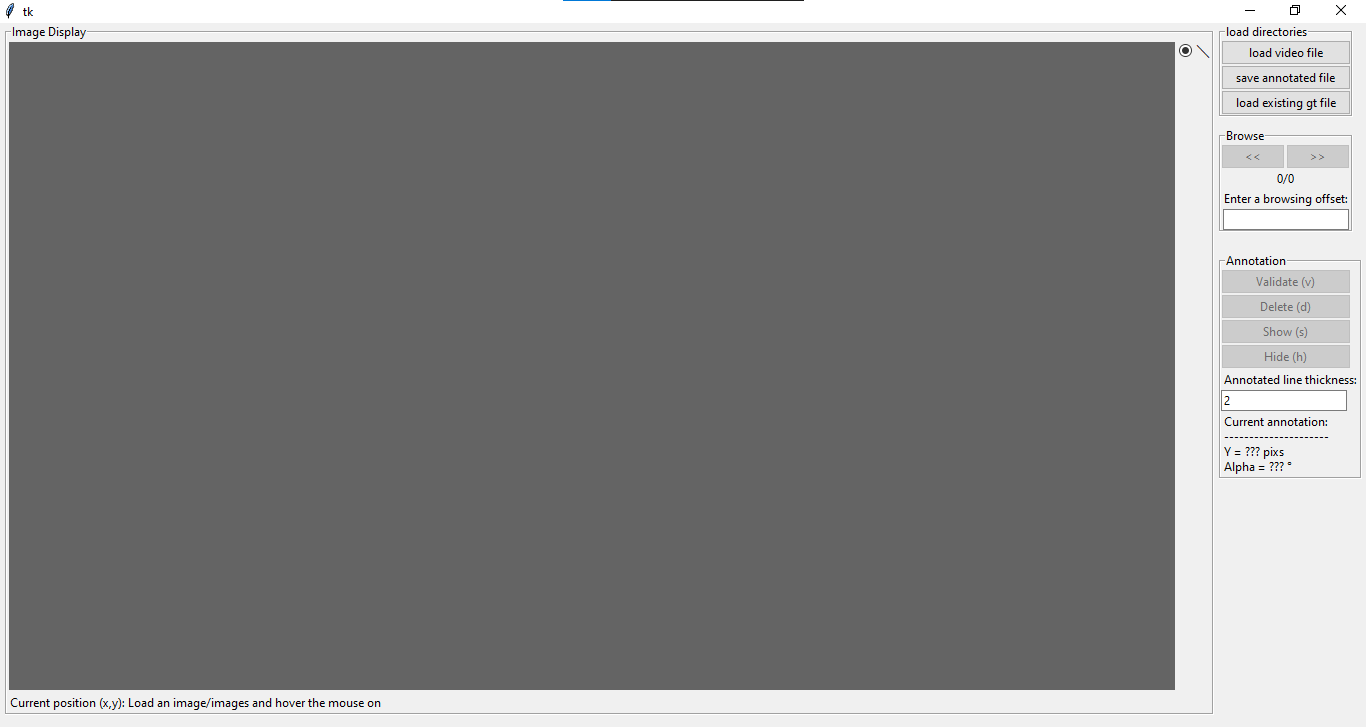}
	\caption{The GUI before video loading}
	\label{fig_overview}
\end{figure*}

Figure~\ref{fig_overview} shows the first interface that pops-up when running the application. It is subdivided into four blocks: image display, load directories, browse, and annotation. This Section discusses the full details of each of mentioned blocks.

\subsection{Load directories block}
\begin{figure}[h!]
	\centering
	\includegraphics[width = 0.7\linewidth, keepaspectratio]{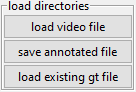}
	\caption{The \textit{Load directories} block}
	\label{fig_load_block}
\end{figure}

\begin{figure}[h!]
	\centering
	\subfigure[]{
		\includegraphics[width = 1\linewidth, keepaspectratio]{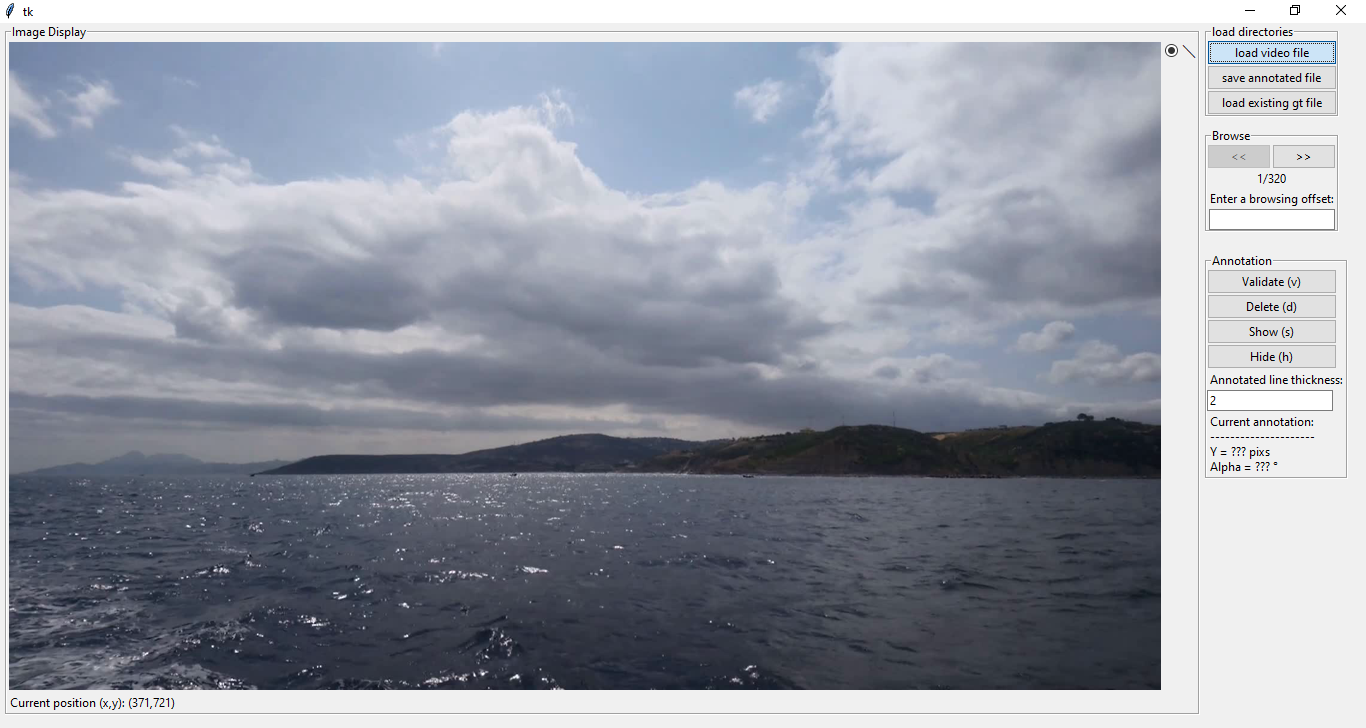}
		\label{fig_loaded_video}}
	\subfigure[]{
		\includegraphics[width = 1\linewidth, keepaspectratio]{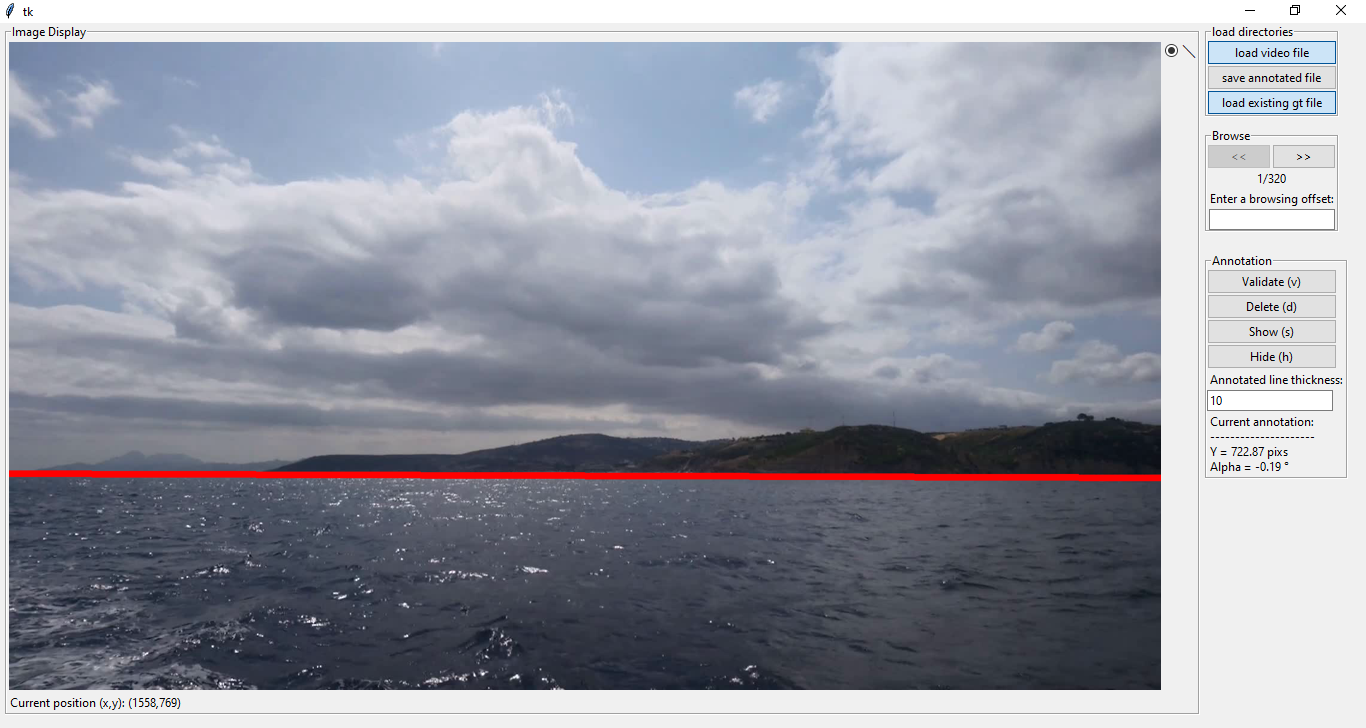}
		\label{fig_loaded_gt_file}}
	\caption{The GUI after loading (a) a video file and an (b) existing annotation file.}
	\label{fig_loaded_video_and_file}
\end{figure}

\textit{Load directories} is the first block to use. As shown in Figure~\ref{fig_load_block}, the block contains three self-explanatory buttons: \textit{load video file}, \textit{save annotated file}, and \textit{load existing gt file}. The software supports two video file formats: \textit{.avi} and \textit{.mp4}. The loaded video will appear on the \textit{Image display block}, which we detail in Section~\ref{sec_imgdisplay}. The third button (\textit{load existing gt file}) allows the user to start the annotation process from an existing annotated file (also known as the Ground Truth file). As the user browses through the video frames, the annotated line of the viewed frame will be drawn in red, as shown in Figure~\ref{fig_loaded_video_and_file}. Such a feature allows the user to review the annotated frames and correct the eventual mistakes. Whether the user annotates the video from scratch or starts the annotation from an existing GT (Ground Truth) file, clicking on the \textit{save annotated file} will always save in the directory specified by the user an annotation file by suffixing the current video file name with \textit{LineGT.npy}.

\subsection{Image display block}
\label{sec_imgdisplay}
\begin{figure}[h!]
	\centering
	\subfigure[]{
		\includegraphics[width = 1\linewidth, keepaspectratio]{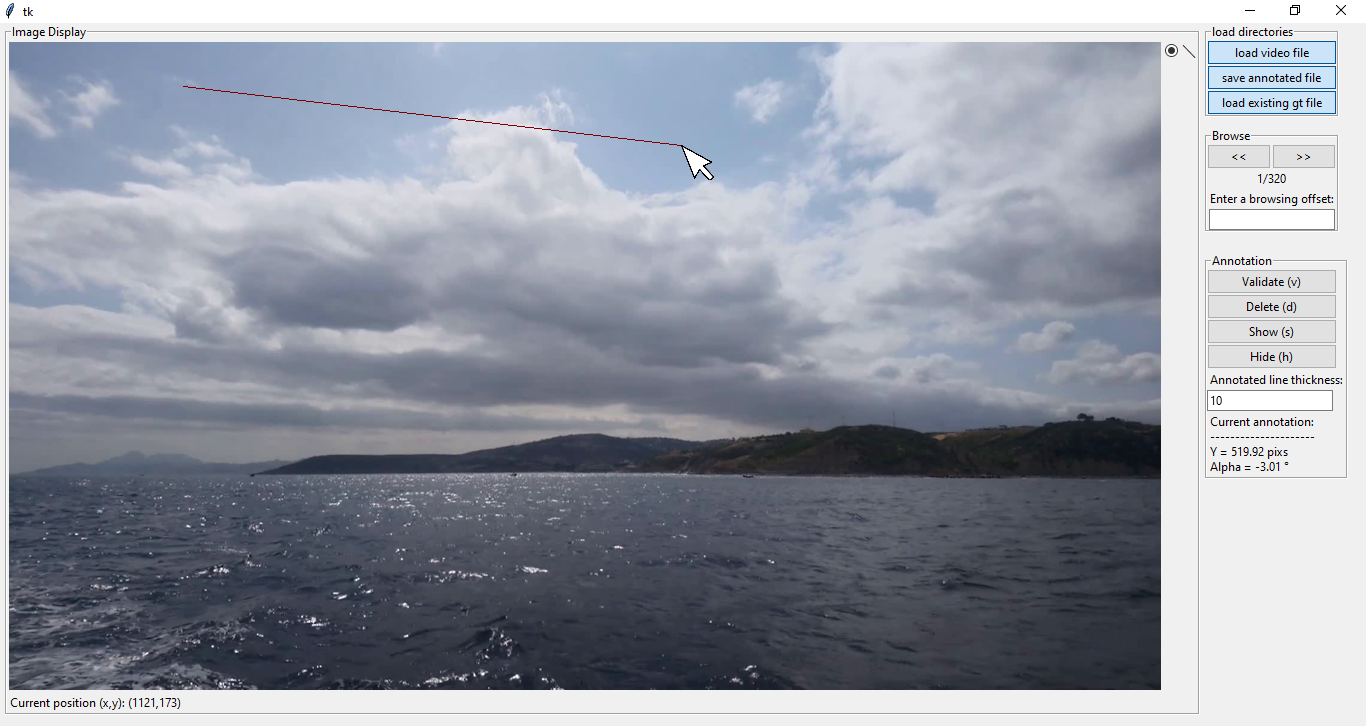}
		\label{fig_drawing_line}}
	\subfigure[]{
		\includegraphics[width = 1\linewidth, keepaspectratio]{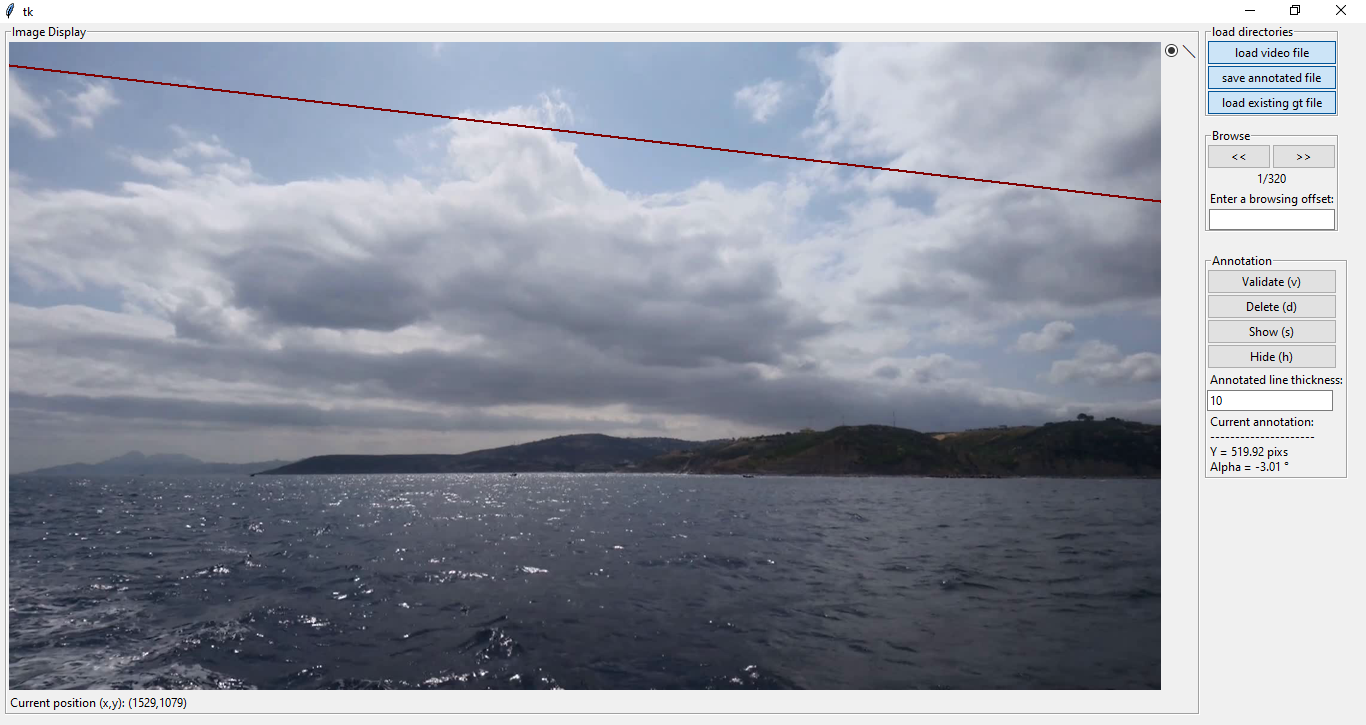}
		\label{fig_finished_line}}
	\caption{The process of line drawing: (a) before releasing the left mouse button; (b) after releasing the left mouse button; (c) the mouse position coordinates.}
	\label{fig_drawn_line}
\end{figure}

\begin{figure}[h!]
	\centering
	\includegraphics[width = 1\linewidth, keepaspectratio]{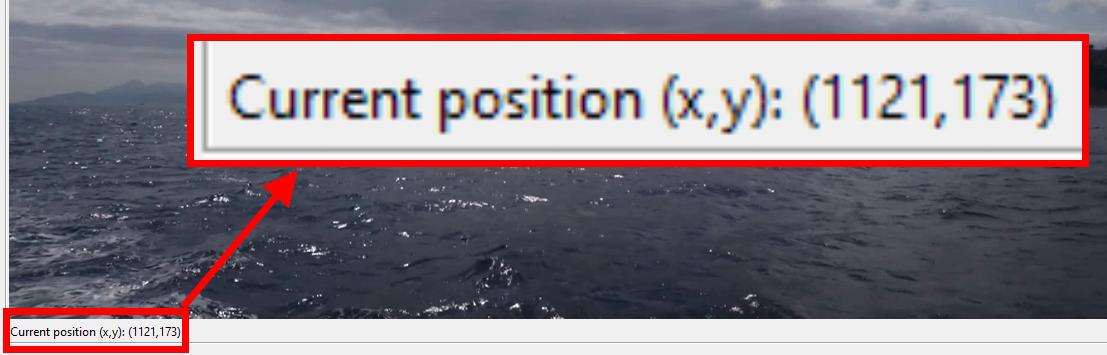}
	\caption{Cartesian coordinates $(x, y)$ of the current mouse position}
	\label{fig_mouse_coordinates}
\end{figure}

The image display block will show the first frame of the loaded video file. The frame size will not change as long as the entire GUI fits into the computer screen. Otherwise, the frame will be down-scaled to the size that would allow the entire GUI to fit onto the computer screen. If the current frame corresponds to an annotated line, that line will be automatically drawn and shown on the displayed frame. The user can draw the annotation line by maintaining the left mouse button. one that button is released, the application will automatically infer and draw the full line on the image display, as shown in Figure~\ref{fig_drawn_line}. If the user is not satisfied with the line he is drawing, i.e., before releasing the left button of the mouse, he can abort the drawing process by right-clicking the mouse before releasing the left mouse button. While the mouse is on the image display, its Cartesian Coordinates $(x, y)$ are constantly updated and displayed on the bottom left of the Image display block, as shown in Figure~\ref{fig_mouse_coordinates}. We note that the image frame displayed in Figure~\ref{fig_drawn_line} is downsized so that the GUI fits in the screen.

Even in such a case, the Cartesian coordinates displayed are computed to correspond to the original size of the frame. This choice is not arbitrary for two reasons: (1) the downsizing factor changes according to the frame size and the computer screen; (2) the position of the annotated line must correspond to the original frame size.
\subsection{Annotation block}
\begin{figure}[h!]
	\centering
	\includegraphics[width = 0.7\linewidth, keepaspectratio]{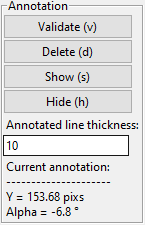}
	\caption{The \textit{Annotation} block}
	\label{fig_annotation_block}
\end{figure}

We show the annotation block in Figure~\ref{fig_annotation_block}. The features of this block involve the manipulation of annotated lines. The \textit{Validate} button validates the non-aborted line drawn, such as the line we mentioned in Figure~\ref{fig_drawn_line}. The \textit{Delete} button deletes the annotated line of the current frame. The \textit{Hide} button hides the annotation of the current frame. It can be shown again using the \textit{Show} button. Instead of using these buttons, the user may prefer keyboard keys. The key corresponding to each button is written between parenthesis (e.g., (v) for \textit{Validate} button). Section~\ref{sec_keymaps} provides a keymap containing the full list of the GUI shortcuts. 

The \textit{Annotated line thickness} shown under the four buttons in Figure~\ref{fig_annotation_block} is a text entry that allows the user to specify the thickness of the annotated line. We note that such a thickness is not an annotation information and serves only a visualization purpose. This entry has built-in features to handle non-conforming thickness values, such as texts or negative numbers. The last element of the \textit{annotation block} is the \textit{Current annotation}, which displays the position and tilt of the annotated line. When no annotation line exists for the current frame, the values displayed will be replaced by three interrogation marks ???.

\subsection{Browse block}
\begin{figure}[h!]
	\centering
	\includegraphics[width = 0.7\linewidth, keepaspectratio]{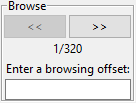}
	\caption{The \textit{Browse} block}
	\label{fig_browse_block}
\end{figure}

After loading a video file, drawing a line, and validating it using the \textit{Validate Button} or its corresponding shortcuts, the user will browse to the next frames. The \textit{Browse block} and its features ensure easy browsing through the video frames. As we show in Figure~\ref{fig_browse_block}, the two buttons $<<$ and $>>$ browse to the previous and next frames, respectively. Easier browsing alternatives are listed in Section~\ref{sec_keymaps}.
Right under the two browsing buttons, the interface shows the index of the current frame and the total number of frames. The browsing offset defaults to 1. This means that the index of the next (or previous) frame is equal to the index of the current frame + 1 (or - 1). Browsing with an offset of 1 slows down the browsing process when the video file has a large number of frames. Thus, the user can quickly go to any video frame by changing the browsing offset in the \textit{Enter a browsing offset} entry.

\section{Annotation files}
\begin{figure}[h!]
	\centering
	\includegraphics[width = 1\linewidth, keepaspectratio]{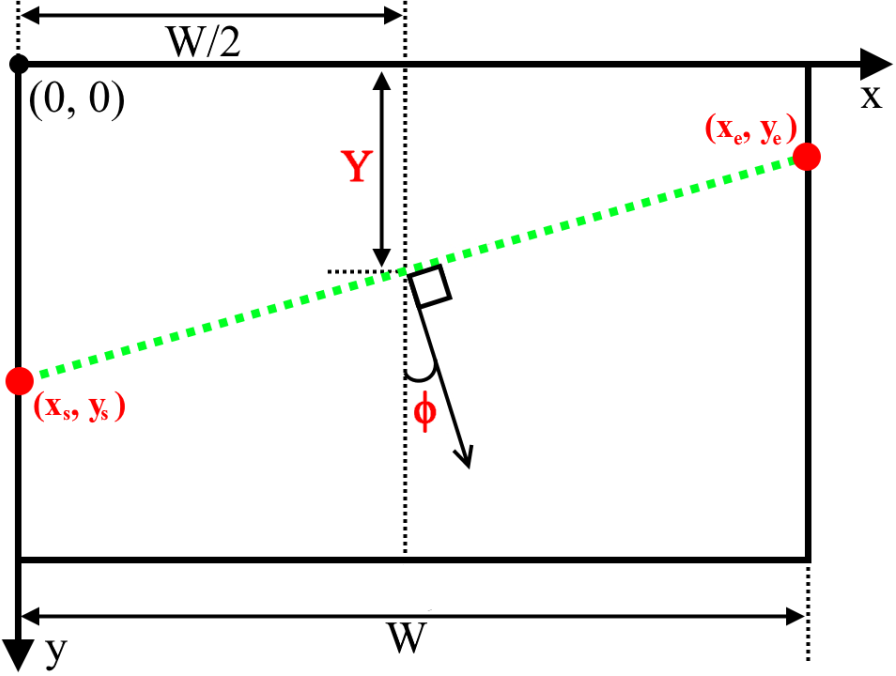}
	\caption{The line parameters saved in the annotation files}
	\label{fig_pars_annotation_file}
\end{figure}

The annotation files take the \textit{.npy} format, which can be loaded on Python using the \textit{Numpy} package~\cite{numpy}. Each \textit{.npy} file will contain $N$ rows, where $N$ is the number of frames in the corresponding video file. We note that even if the user does not annotate all the frames, the number of rows in the saved \textit{.npy} will always be $N$. Each row contains five columns, and each column is a scalar number representing the following parameters: $Y$, $\phi$, $x_s$, $x_e$, $y_s$, $y_e$, where $Y$ is the line position in pixels, $\phi$ is the line tilt in
degrees, $(x_s, y_s)$ are Cartesian coordinates of the starting point, and $(x_e, y_e)$ are Cartesian coordinates of the ending point. We visually illustrate these five parameters in Figure~\ref{fig_pars_annotation_file}. We note that $\phi$ is zero when the line is horizontal and increases in the anti-clockwise rotation; for instance, the line in Figure~\ref{fig_pars_annotation_file} has a positive tilt value $(\phi > 0)$. We note that non-annotated frames will correspond to rows where scalars $Y$, $\phi$, $x_s$, $x_e$, $y_s$, and $y_e$ will be \textit{np.nan}, which is the \textit{Numpy} object for \textit{Not A Number}.

\section{Keymap for quick and easy annotation}
\label{sec_keymaps}
This Section lists all the shortcut keys we chose to quickly and easily annotate a video file. These keys allow the user to annotate an entire video file with minimal hand movement. For instance, the keymap in Table~\ref{tab_keymap} indicates that the user can annotate the entire video file using the mouse only: draw a line with the left mouse button, validate that line with the right mouse button, and browse to the next frame by rotating the mouse wheel instead of clicking on the next button ($>>$) shown in Figure~\ref{fig_browse_block}. Another significant productivity feature in our software is the annotation replication, which can be done using the \textit{w} key. For instance, let’s assume that the camera does not move for a given period, which is very likely to happen for cameras on non-moving platforms such as shore pylons. The SMD~\cite{EOsurvey2017} contains more than 17,222 video frames captured with such a camera set-up. If the period we just mentioned lasts, for instance, only five seconds, there would be $150$ frames\footnote{$5 \times$ Frames per seconds = $5 \times 30$} where the horizon's position and orientation is the same (because the camera's position and orientation is the same). In other words, all $150$ frames we mentioned will have the same annotation. In such a case, the user should browse to the $150$-th non-annotated frame, annotate it, and then click on the \textit{w} key to replicate that annotation on all the previous $149$ non-annotated frames. This is much easier and faster than drawing and validating a line for each of the $150$ non-annotated frames. Table~\ref{tab_keymap} shows the complete list of keys facilitating the annotation process. 

\begin{table}[!h] 
	\centering
	\caption{\label{tab_keymap}\footnotesize{Keymap of the GUI}}\vspace{-2mm}
	\setlength{\extrarowheight}{5pt}
	{\footnotesize
		\begin{tabular}{|
				P{0.2\linewidth}|
				P{0.71\linewidth}|}
			\hline
			Keys&Actions\\
			\hline
			Mouse wheel (roll-up)&Browse to the next frame 
			\\
			\hline
			Right arrow key ($\rightarrow$) &Browse to the next frame
			\\
			\hline
			Mouse wheel (roll-down)&Browse to the previous frame
			\\
			\hline
			Left arrow key ($\leftarrow$) & Browse to the previous frame
			\\
			\hline
			v&Validate the drawn line
			\\
			\hline
			Left mouse button&Validate the drawn line
			\\
			\hline
			w&Replicate the annotation of the current frame on all previous \textit{non-annotated} frames
			\\
			\hline
			s&Show the annotated line (if it exists)
			\\
			\hline
			
			h&Hide the annotated line (if it exists)
			\\
			\hline
			d&Delete the annotated line
			\\
			\hline
			Enter key&Validates entered texts (e.g., \textit{Annotated line thickness} in Figure~\ref{fig_annotation_block})
			\\
			\hline
		\end{tabular}
	}
\end{table}

\section{Message boxes}
\begin{figure}[h!]
	\centering
	\subfigure[]{
		\includegraphics[width = 1\linewidth, keepaspectratio]{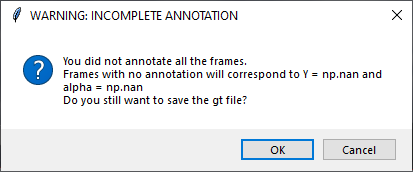}
		\label{fig_warning_1}}
	\subfigure[]{
		\includegraphics[width = 1\linewidth, keepaspectratio]{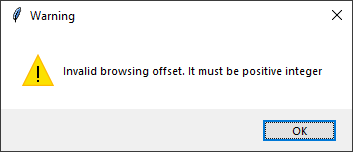}
		\label{fig_warning_2}}
	\caption{Two instances of message boxes}
	\label{fig_warnings}
\end{figure}

The GUI includes several message boxes to guide the user or warn him with relevant information. Figure~\ref{fig_warning_1} shows message box examples that would pop-up when the user tries to save an annotation file with missing annotations or enters an invalid browsing offset.

\section{Conclusion}
We proposed in this paper a horizon line annotation software and justified its importance in several aspects, which include the lack of current public datasets in terms of more maritime conditions as well as the mistakes in their annotation files, the preference of researchers to collect an annotate their own maritime images, and the absence of a free public software dedicated to the horizon annotation task. We discussed and described all blocks of the graphical user interface. More importantly, we provided a complete keymap list that corresponds to several productivity features we designed to quickly and easily annotate the video frames. We provided the software as one file that would install all required files, including the executable application \textit{Horizon Annotator.exe}.

\printbibliography

\end{document}